\documentclass[journal]{IEEEtran}
\usepackage{cite}
\usepackage{times}
\usepackage{epsfig}
\usepackage{graphicx}
\usepackage{amsmath}
\usepackage{amssymb}
\usepackage{algorithm, algorithmic}
\usepackage{enumerate}
\usepackage{multirow}
\usepackage{xcolor}
\usepackage{graphics}
\usepackage{epsfig}
\usepackage{epstopdf}
\usepackage{subfigure}
\usepackage[switch]{lineno}
\usepackage{caption}
\usepackage{changepage}
\usepackage{color}
\usepackage{url}

\newcommand{\tabincell}[2]{\begin{tabular}{@{}#1@{}}#2\end{tabular}}

\begin{document}

\title{{Robust Lane Detection from Continuous Driving Scenes Using Deep Neural Networks}}

\author{Qin~Zou,
        Hanwen~Jiang,
        Qiyu~Dai,
        Yuanhao~Yue,
        Long~Chen,
        and~Qian Wang \\
        $\ $ \\
        {\color{blue}\url{https://github.com/qinnzou/Robust-Lane-Detection}}
\thanks{This research was supported by the National Natural Science
Foundation of China under grant 61872277, 61773414 and 41571437, and the Hubei Provincial Natural
Science Foundation under grant 2018CFB482.
}
\thanks{Q.~Zou, Y.~Yue and Q.~Wang are with the
School of Computer Science, Wuhan University, Wuhan 430072,
P.R.~China (E-mails: \{qzou, yhyue, qianwang\}@whu.edu.cn).}
\thanks {H.~Jiang is
with the Department of Electronic Information, Wuhan University, Wuhan 430072, P.R.~China (E-mail:
hwjiang@whu.edu.cn). }
\thanks {Q.~Dai is
with the Department of Power and Mechanical Engineering, Wuhan University, Wuhan 430072, P.R.~China (E-mail:
qiyudai@whu.edu.cn). }
\thanks{L.~Chen is with the School of Data and Computer Science, Sun Yat-Sen University, Guangzhou 518001,
P.R.~China (E-mail: chenl46@mail.sysu.edu.cn).}

}

\markboth{IEEE Transactions on Vehicular Technology, 2019}
{Shell \MakeLowercase{\textit{et al.}}: }

\maketitle


\begin{abstract}
Lane detection in driving scenes is an important module for autonomous vehicles and advanced driver assistance systems. In recent years, many sophisticated lane detection methods have been proposed. However, most methods focus on detecting the lane from one single image, and often lead to unsatisfactory performance in handling some extremely-bad situations such as heavy shadow, severe mark degradation, serious vehicle occlusion, and so on. In fact, lanes are continuous line structures on the road. Consequently, the lane that cannot be accurately detected in one current frame may potentially be inferred out by incorporating information of previous frames. To this end, we investigate lane detection by using multiple frames of a continuous driving scene, and propose a hybrid deep architecture by combining the convolutional neural network (CNN) and the recurrent neural network (RNN). Specifically, information of each frame is abstracted by a CNN block, and the CNN features of multiple continuous frames, holding the property of time-series, are then fed into the RNN block for feature learning and lane prediction. Extensive experiments on two large-scale datasets demonstrate that, the proposed method outperforms the competing methods in lane detection, especially in handling difficult situations. 
\end{abstract}

\begin{IEEEkeywords}
Convolutional neural network, LSTM, lane detection,
semantic segmentation, autonomous driving.
\end{IEEEkeywords}

\IEEEpeerreviewmaketitle

\section{Introduction}

\IEEEPARstart{W}{ith} the rapid development of high-precision optic sensors and electronic sensors, high-efficient and high-effective computer vision and machine learning algorithms, real-time driving scene understanding has become more and more realistic to us. Many research groups from both academia and industry have invested large amount of resources to develop advanced algorithms for driving scene understanding, targeting at either an autonomous vehicle or an advanced driver assistance system (ADAS). Among various research topics of driving scene understanding, lane detection is a most basic one. Once lane positions are obtained, the vehicle will know where to go, and avoid the risk of running into other lanes~\cite{wang2018learning}.

\begin{figure}
	\centering
	\includegraphics[width=0.95\linewidth]{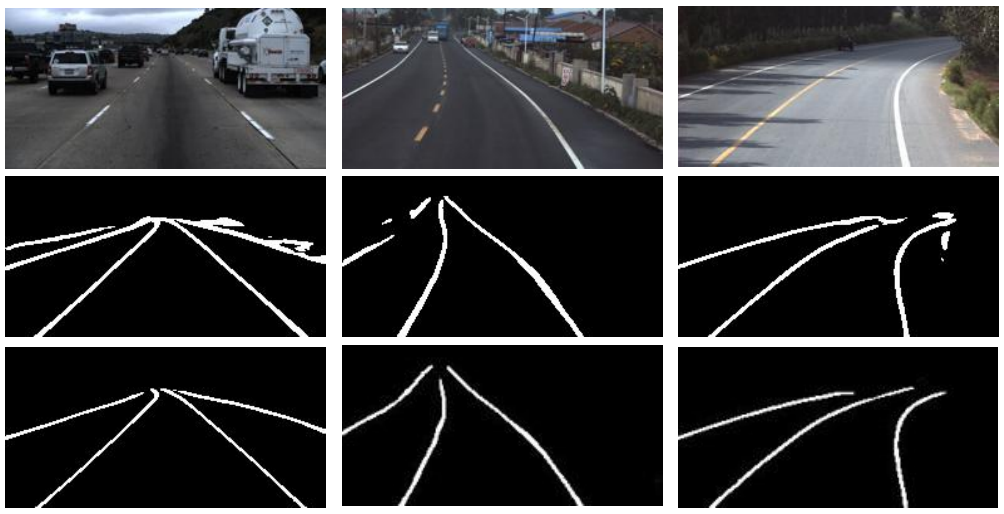}
	\caption{{{Lane detection in challenging situations. Top row: three example images of different driving scenes. Middle row: lane detection using only the current frame. Bottom row: lane detection using four previous frames and the current frame with the proposed method.}}}
	\label{fig:sample3}
\end{figure}

In recent years, a number of lane-detection methods have been proposed with sophisticated performance as reported in the literatures. Among these methods, some represent the lane structure with geometry models~\cite{Wang2004LaneDU,Borkar2011}, some formulate lane detection as energy minimization problems~\cite{Wojek2008eccv,Hur2013MultilaneDI}, and some segment the lane by using supervised learning strategies~\cite{Kim2017FastLM,Li2017DeepNN,Lee2017VPGNetVP,Huang2018SpatialTemproalBL}, and so on. However, most of these methods limit their solutions by detecting road lanes in one current frame of the driving scene, which would lead to low performance in handling challenging driving scenarios such as heavy shadows, severe road mark degradation, serious vehicle occlusion, as shown in the top three images in Fig.~\ref{fig:sample3}. In these situations, the lane may be predicted with false direction, or may be detected in partial, or even cannot be detected at all. One main reason is that, the information provided by the current frame is not enough for accurate lane detection or prediction.

Generally, road lanes are line structures and continuous on the pavement, appeared either in solid or dash. As the driving scenes are continuous and are largely overlapped between two neighboring frames, the position of lanes in the neighboring frames are highly related. More precisely, the lane in the current frame can be predicted by using multiple previous frames, even though the lane may suffer from damage or degradation brought by shadows, stains and occlusion. This motivates us to study lane detection by using images of a continuous driving scene.

Meanwhile, we notice that deep learning as an emerging technology has demonstrated state-of-the-art, human-competitive, and sometimes better-than-human performance in solving many computer vision problems such as object detection~\cite{Girshick2015FastR,Redmon2016YouOL,He2017MaskR}, image classification/retrieval~\cite{vgg2015,tip2019road,zou19tmm} and semantic segmentation~\cite{Shelhamer2015FullyCN,Ronneberger2015UNetCN,Badrinarayanan2017SegNetAD,Zou2018deepcrack}. There are mainly two types of deep neural networks. One is the deep convolutional neural network (DCNN), which often processes the input signal with several stages of convolution, and is talent in feature abstraction for images and videos. The other is the deep recurrent neural network (DRNN), which recursively processes the input signal by separating it into successive blocks and building full connection layers between them for status propagation, and is talent in information prediction for time-series signals. Considering the problem of lane detection, the continuous driving scene images are apparently time-series, which is appropriate to be processed with DRNN. However, if the raw image of a frame is taken as the input, for example an 800$\times$600 image, then the dimension of the vector is 480,000,  which is intolerable as a full-connection layer by a DRNN network, as regarding to the heavy computational burden on a tremendous number of weight parameters it would bring about. In order to reduce the dimensionality of the input data, and meanwhile to maintain enough information for lane detection, the DCNN can be selected as a feature extractor to abstract each single image.

Based on the discussion above, a hybrid deep neural network is proposed for lane detection by using multiple continuous driving scene images. The proposed hybrid deep neural network combines the DCNN and the DRNN. In a global perspective, the proposed network is a DCNN, which takes multiple frames as an input, and predict the lane of the current frame in a semantic-segmentation manner. A fully convolution DCNN architecture is presented to achieve the segmentation goal. It contains an encoder network and a decoder network, which guarantees that the final output map has the same size as the input image. In a local perspective, features abstracted by the encoder network of DCNN are further processed by a DRNN. A long short-term memory (LSTM) network is employed to handle the time-series of encoded features. The output of DRNN is supposed to have fused the information of the continuous input frames, and is fed into the decoder network of the DCNN to help predict the lanes.

The main contributions of this paper lie in three-fold:
\begin{enumerate}[    $\vcenter{\hbox{\tiny$\bullet$}}$]
	
	\item First, for the problem that lane cannot be accurately detected using one single image in the situation of shadow, road mask degradation and vehicle occlusion, a novel method that using continuous driving scene images for lane detection is proposed. As more information can be extracted from multiple continuous images than from one current image, the proposed method can more accurately predicted the lane comparing to the single-image-based methods, especially in handling the above mentioned challenging situations.
	
	\item Second, for seamlessly integrating the DRNN with DCNN, a novel fusion strategy is presented, where the DCNN consists of an encoder and a decoder with fully-convolution layers, and the DRNN is implemented as an LSTM network.  The DCNN abstracts each frame into a low-dimension feature map, and the LSTM takes each feature map as a full-connection layer in the time line and recursively predicts the lane. The LSTM is found to be very effective for information prediction, and significantly improve the performance of lane detection in the semantic-segmentation framework.
	
	\item Third, two new datasets are collected for performance evaluation. One dataset collected on 12 situations with each situation containing hundreds of samples. The other dataset are collected on rural roads, containing thousands of samples. These datasets can be used for quantitative evaluation for different lane-detection methods, which would help promote the research and development of autonomous driving. Meanwhile, another dataset, the TuSimple dataset, is also expanded by labeling more frames.
\end{enumerate}

The remainder of this paper is organized as follows.
Section~\ref{sec:relate} reviews the related work.
Section~\ref{sec:method} introduces the proposed hybrid deep neural network, including deep convolutional neural network,
deep recurrent neural network, and training strategies.
Section~\ref{sec:experiment} reports the experiments and results.
Section~\ref{sec:conclusion} concludes our work and briefly discusses the possible future work.

\section{Related work}\label{sec:relate}
\subsection{Lane Detection Methods}
In the past two decades, a number of researches have been made in the fields of lane detection and prediction~\cite{chen2010iv,Hillel2014,Yenikaya2013,Li2014ASD}. These methods can be roughly classified into two categories: traditional methods and deep learning based methods. In this section, we briefly overview each category.

{\textbf{Traditional methods.}} Before the advent of deep learning technology, road lane detection were geometrically modeled in terms of line detection or line fitting, where primitive information such as gradient, color and texture were used and energy minimization algorithms were often employed.

{\it 1) Geometric modelling.} Most methods in this group follow a two-step solution, with edge detection at first and line fitting at second~\cite{Aly2008RealTD,McCall2006VideobasedLE,Zhou2010ANL,Borkar2011}. For edge detection, various types of gradient filters are employed. In~\cite{Aly2008RealTD}, edge features of lanes in bird-view images were calculated with Gaussian filter. In~\cite{McCall2006VideobasedLE} and \cite{Zhou2010ANL}, Steerable filter and Gabor filter~\cite{Zhou2010ANL} are also investigated for lane-edge detection, respectively.  In~\cite{selver16}, for rail extraction from videos, the image was pre-processed with Gabor filtering with kernels in different directions, and then post-processed with morphology operates to link the gaps and remove unwanted blocks. Beside gradient, color and texture were also studied for lane detection or segmentation. In~\cite{Choi2010IlluminationIL}, template matching was performed to find lane candidates at first, and then lanes were extracted by color clustering, where the task of illumination-invariant color recognition was assigned to a multilayer perceptron network.  In~\cite{He2004ColorbasedRD}, color was also used for road lane extraction.

For line fitting, Hough Transformation (HT) models were often exploited, e.g., polar randomized HT~\cite{Borkar2011}. Meanwhile, curve-line fitting can be used either as a tool to post-process the HT results or to replace HT directly. For example, in~\cite{Wang1998LaneDU}, Catmull-Rom Spline was used to construct smooth lane by using six control points, in~\cite{Wang2004LaneDU,Deng2013ARS}, B-snake was employed for curve fitting for lane detection. In addition, the geometry modeling-based lane detection can also be implemented with stereo visions~\cite{Caraffi2007OffRoadPA,Wedel2009BSplineMO,Danescu2009ProbabilisticLT}, in which the distance of the lane can be estimated.

{\it 2) Energy minimization.} As an energy-minimization model, conditional random field (CRF) is often used for solving multiple association tasks, and is widely applied in traffic scene understanding~\cite{Wojek2008eccv}. In~\cite{Hur2013MultilaneDI}, CRF was used to detect multiple lanes by building an optimal association of multiple lane marks in complex scenes. With all defined unary and clique potentials, the energy of the graph can be obtained and optimized by energy minimization. Energy minimization can also be embedded in searching optimal modeling results in lane fitting. In~\cite{Kang1996iv}, an active line model was developed, which constructed the energy function with two parts, i.e., an external part - the normalized sum of image gradients along the line, and an internal part - the continuity of two
neighboring lines.

For lane tracking in continue frames, Kalman filter is widely used~\cite{Borkar2009RobustLD,Suttorp2006LearningOK,Mammeri2014LaneDA}. The Kalman filter works well in locating the position of lane markings and estimating the curvature of lanes with some state vectors. Another widely-used method for lane tracking is the particle filter, which also has the ability to track multiple lanes~\cite{Linarth2011OnFT}. In~\cite{Loose2009KalmanPF}, these two filters were combined together as a Kalman-Particle filter, which was reported to achieve more stable performance in simultaneous tracking of multiple lanes. In slice-by-slice medical image processing, information of previous images was found to be helpful for successive images. In~\cite{selver2014ct}, considering that adjacent images have similar object sizes and shapes, the author used a previously segmented image to create a region of interest in the upcoming one, and obtained improved results.

{\textbf{Deep-learning-based methods.}} The research of lane detection leaps into a new stage with the development and successful application of deep learning. A number of deep learning based lane detection methods have been proposed in the past several years. We categorize these methods into four groups.

{\it 1) Encoder-decoder CNN.} The encoder-decoder CNN is typically used for semantic segmentation~\cite{Badrinarayanan2017SegNetAD}. In~\cite{Kim2017EndToEndEL}, lane detection was investigated in a transfer learning framework. The end-to-end encoder-decoder network is built on the basis of road scene object segmentation task, trained on ImageNet. In~\cite{Neven2018TowardsEL}, a network called LaneNet was proposed. LaneNet is built on SegNet, but has two decoders. One decoder is a segmentation branch, detecting lanes in a binary mask. The other is an embedding branch, segmenting the road. As the output of the network is generally a feature map, clustering and curve-fitting algorithms were then required to produce the final results In~\cite{BrulsMM018icra}, real-time road marking segmentation was exploited in a condition of lack of large amount of labeled data for training. A novel weakly-supervised strategy was proposed, which utilized additional sensor modalities to generate large quantities of annotated images. These annotated images then enabled the training of a U-Net inspired network and achieved real-time and accurate road marking detection.

{\it 2) FCN with optimization algorithms.} Fully-convolutional neural networks (FCN) with optimization algorithms are also widely used for lane detection. In~\cite{Huval2015AnEE}, sliding window with overfeat features was applied for parsing the driving scene, where the lane is detected as a separate class. In~\cite{Kim2017FastLM}, CNN with fully-connective layers was used as a primitive decoder, in which hat-shape kernels were used to infer the lane edges and the RANSAC was applied to refine the results. This model demonstrated enhanced performance when implemented in an extreme-learning framework.

In~\cite{He2016AccurateAR}, a dual-view CNN (DVCNN) framework was proposed to process top-view and front-view images. A weighted hat-like filter was applied to the top view to find lane candidates, which were then processed by a CNN. An enhance version of this method was proposed in~\cite{He2016LaneMD}, where CNN was used to process the data produced by point cloud registration, road surface segmentation and orthogonal projection. In~\cite{Gurghian2016DeepLanesEL}, algorithms for generating semi-artificial images were presented, and lane detection was achieved by using a CNN with fully-connected layer and softmax classification. On the purpose of multi-task learning, a VPGNet was proposed in~\cite{Lee2017VPGNetVP}, which has a shared feature extractor and four similar branch layers. Optimization algorithms such as clustering and subsampling are followed to achieve the goal of lane detection, lane-marking identification and vanishing-point extraction. Similar methods were also presented in~\cite{Bailo2017RobustRM}. In~\cite{Pan2018SpatialAD}, a spatial CNN was proposed to improve the performance of CNN in detecting long continuous shape structures. In this method, traditional layer-by-layer convolutions were generalized into slice-by-slice convolutions within the feature maps, which enables message passing between pixels across rows and columns. In~\cite{Huang2018SpatialTemproalBL}, spatial and temporal constraints were also considered for estimating the position of the lane.

{\it 3) `CNN+RNN'.} Considering that road lane is continuous on the pavement, a method combining CNN and RNN was presented in~\cite{Li2017DeepNN}. In this method, a road image was firstly divided into a number of continuous slices. Then, a convolutional neural network was employed as feature extractor to process each slice. Finally, a recurrent neural network was used to infer the lane from feature maps obtained on the image slices. This method was reported to get better results than using CNN only. However, the RNN in this method can only model time-series feature in a single image, and the RNN and CNN are two separated blocks.

\begin{figure*}
	\centering
	\includegraphics[width=1.0\linewidth]{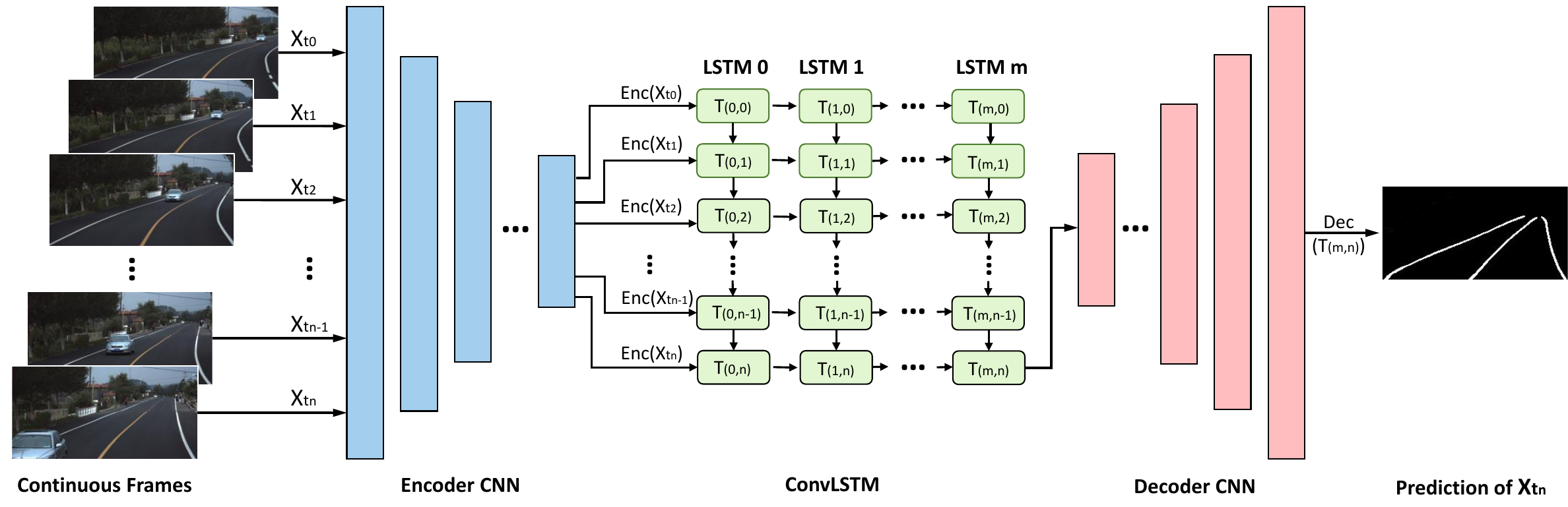}
	\caption{{{Architecture of the proposed network.}}}
	\label{fig:arch-all}
\end{figure*}

{\it 4) GAN model.} The generative adversarial network (GAN)~\cite{Goodfellow2014GenerativeAN}, which consists of a generator and a discriminator, is also employed for lane detection~\cite{Ghafoorian2018ELGANEL}. In this method, an embedding-loss GAN (EL-GAN) was proposed for semantic segmentation of driving scenes, where the lanes were predicted by a generator based on the input image, and judged by a discriminator with shared weights. The advantage of this model is that, the predicted lanes are thin and accurate, avoiding marking large soft boundaries as usually brought by CNNs.

Different with the above deep-learning-based methods, the proposed method models lane detection as a time-series problem, and detects lanes in a number of continuous frames other than in only one current frame. With richer information, the proposed method can obtain a robust performance in lane detection in challenging conditions. Meanwhile, the proposed method seamlessly integrates the CNN and RNN, which provides an end-to-end trainable network for lane detection.

\subsection{ConvLSTM for Video Analysis}
LSTM is a basic building block in deep learning, which is talent in processing temporal information. Convolutional LSTM (ConvLSTM) is a kind of LSTM equipped with convolution operations, and has been widely used in video analysis due to their feedback mechanism on temporal dynamics and the abstraction power on image representation~\cite{Shi2015ConvolutionalLN}. There are different ways to use ConvLSTM as a basic building block. For semantic video segmentation specifically, ConvLSTMs are popularly used for efficiently processing the two-dimensional time-series inputs. In~\cite{2018YouTubeVOSSV}, the video segmentation task was transformed into a regression problem, where a ConvLSTM was set between the convolutional encoder-decoder to predict every frame. In~\cite{Tokmakov2017LearningVO}, a convolutional GRU was introduced to fuse the two-stream representation obtained by a normal convolutional appearance network and an optical-flow motion network. A different structure was presented in~\cite{Zhu2017TORNADOAS}, which placed a two-stream ConvLSTM after the encoder to get spatial and temporal region proposals. In~\cite{Song2018PyramidDD}, a spatio-temporal saliency learning module was constructed by placing a pyramid dilated bidirectional ConvLSTM (PDB-ConvLSTM) after the spatial saliency learning module. The ConvLSTM was also employed in some other video-analysis tasks such as anomaly detection~\cite{medel2016anomaly}, sequence prediction\cite{villegas2017decomposing}, and passenger-demand prediction~\cite{Ke2017ShortTermFO}~\cite{Liang2019ADS}.

\section{Proposed Method}\label{sec:method}
In this section, we introduce a novel hybrid neutral network by fusing DCNN and DRNN together to accomplish the lane-detection task.

\subsection{System Overview}
Lanes are solid- or dash- line structures on the pavement, which can be detected from a geometric-modeling or semantic-segmentation way in one single image. However, these models are not promising to be applied to practical ADAS systems due to their unsatisfying performance under challenging conditions such as heavy shadow, severe mark degradation and serious vehicle occlusion. The information in a single image is found to be insufficient to support a robust lane detection.

The proposed method combines the CNN and RNN for lane detection, with a number of continuous frames of driving scene. Actually, in the continuous driving scene, images captured by automobile cameras are consecutive, and lanes in one frame and the previous frame are commonly overlapped, which enables lane detection in a time-series prediction framework. RNN is adaptive for lane detection and prediction task due to its talent in continuous signal processing, sequential feature extracting and integrating. Meanwhile, CNN is talent in processing large images.
By recursive operations of convolution and pooling, an input image can be abstracted as feature map(s) in a smaller size. These feature maps obtained on continuous frames hold the property of time-series, and can be well handled by an RNN block.

In order to integrate CNN and RNN as an end-to-end trainable network , we construct the network in an encoder-decoder framework. The architecture of the proposed network is shown in Fig.~\ref{fig:arch-all}. The encoder CNN and decoder CNN are two fully convolutional network. With a number of continuous frames as an input, the encoder CNN processes each of them and get a time-series of feature maps. Then the feature maps were input into the LSTM network for lane-information prediction. The output of LSTM is them fed into the decoder CNN to produce a probability map for lane prediction. The lane probability map has the same size of the input image.

\subsection{Network Design}
{\it 1) LSTM network.} Modelling the multiple continuous frames of driving scene as time-series, the RNN block in the proposed 	network accepts feature map extracted on each frame by the encoder CNN as input. To handle various time-series data, different kinds of RNN have been proposed, e.g., LSTM and GRU. In this work, an LSTM network	is employed, which generally outperforms the traditional RNN model with its ability in forgetting unimportant information and remembering essential features, by using cells in the network to judge whether the information is important or not. A double-layer LSTM is applied, with one layer for sequential feature extraction and the other for integration. The traditional full-connection LSTM is time- and computation- consuming. Therefore, we utilize the convolutional LSTM (ConvLSTM)~\cite{Shi2015ConvolutionalLN} in the proposed network. The ConvLSTM replaces the matrix multiplication in every gate of LSTM with convolution operation, which is widely used in end-to-end training and feature extraction from time-series data.

The activations of a general ConvLSTM cell at time \textit{t} can be formulated as
\begin{equation}
\begin{split}
&\mathcal{C}_t \!= \!f_t\circ \mathcal{C}_{t-1} \!+ i_t\circ tanh(W_{xc}\ast \!\mathcal{X}_t + W_{hc}\!\ast \!\mathcal{H}_{t-1} +\!b_c),\\
&f_t = \sigma(W_{xf}\ast \mathcal{X}_t + W_{hf}\ast \mathcal{W}_{t-1} + W_{cf}\circ \mathcal{C}_{t-1} + b_f),\\
&o_t = \sigma(W_{xo}\ast \mathcal{X}_t + W_{ho}\ast \mathcal{W}_{t-1} + W_{co}\circ \mathcal{C}_{t-1} + b_o),\\
&i_t= \sigma(W_{xi}\ast \mathcal{X}_t + W_{hi}\ast \mathcal{W}_{t-1} + W_{ci}\circ \mathcal{C}_{t-1} + b_i),\\
&\mathcal{H}_t = o_t \circ tanh(\mathcal{C}_t),
\end{split}
\end{equation}
where $\mathcal{X}_t$ denotes the input feature maps extracted by the encoder CNN at time $t$. $\mathcal{C}_t$, $\mathcal{H}_t$ and $\mathcal{C}_{t-1}$, $\mathcal{H}_{t-1}$ denote the memory and output activations at time $t$ and $t$-$1$, respectively. $\mathcal{C}_t$, $i_t$, $f_t$ and $o_t$ denote the cell, input, forget and output gates, respectively. $W_{xi}$ is the weight matrix of the input $\mathcal{X}_t$ to the input gate, ${b}_i$ is the bias of the input gate. The meaning of other $W$ and ${b}$ can be inferred from the above rule. $\sigma$($\cdot$) represents the sigmoid operation and $tanh$($\cdot$) represents the hyperbolic tangent non-linearities. `$\ast$' and `$\circ$' denote the convolution operation and the Hadamard product, respectively.

In our network, the input and output size of the ConvLSTM are equal to the size of the feature map produced by the encoder, namely, $8\times16$ for the UNet-ConvLSTM and $4\times8$ for the SegNet-ConvLSTM. The size of the convolutional kernel is $3\times3$.
The ConvLSTM is equipped with 2 hidden layers, and each hidden layer has a dimension of 512.

{\it 2) Encoder-decoder network.} The encoder-decoder framework models lane detection as a semantic-segmentation task. If the architecture of the
encoder-decoder network can support an output with the same size of input, the whole network will be easily trained in an end-to-end manner. In the encoder part, convolution and pooling are used for image abstraction and feature extraction. While in the decoder part, deconvolution and upsampling are used to grasp and highlight the information of targets and spatially reconstruct them.

\begin{figure}[t!]
\centering
\begin{minipage}[b]{0.49\linewidth}
  \centering
  \centerline{\includegraphics[width=4.6cm]{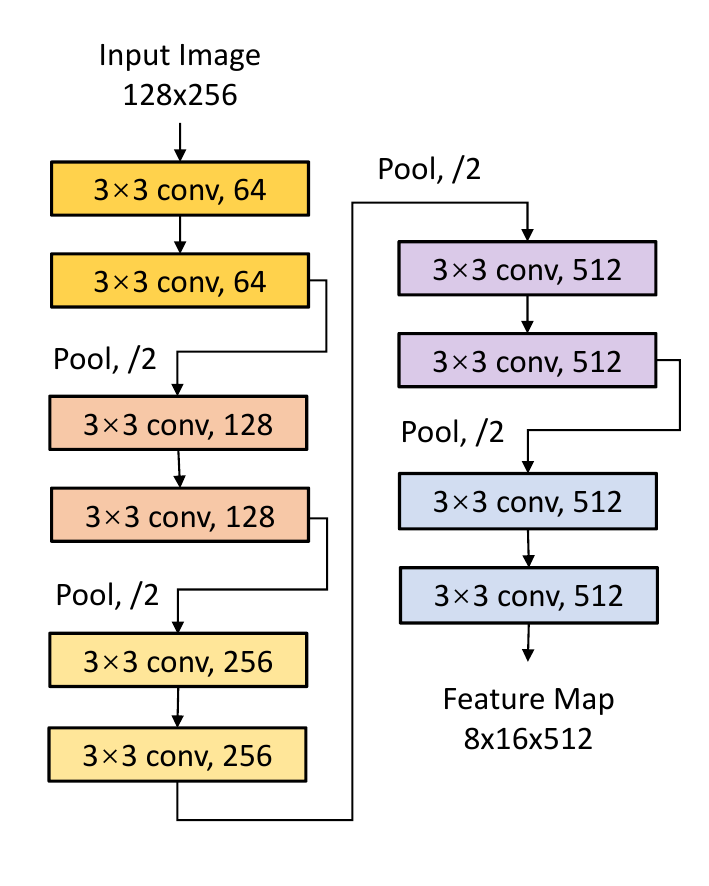}}\vspace{-0mm}
  \centerline{(a)}\medskip
\end{minipage}
\hspace{-1mm}
\begin{minipage}[b]{0.49\linewidth}
  \centering
  \centerline{\includegraphics[width=4.6cm]{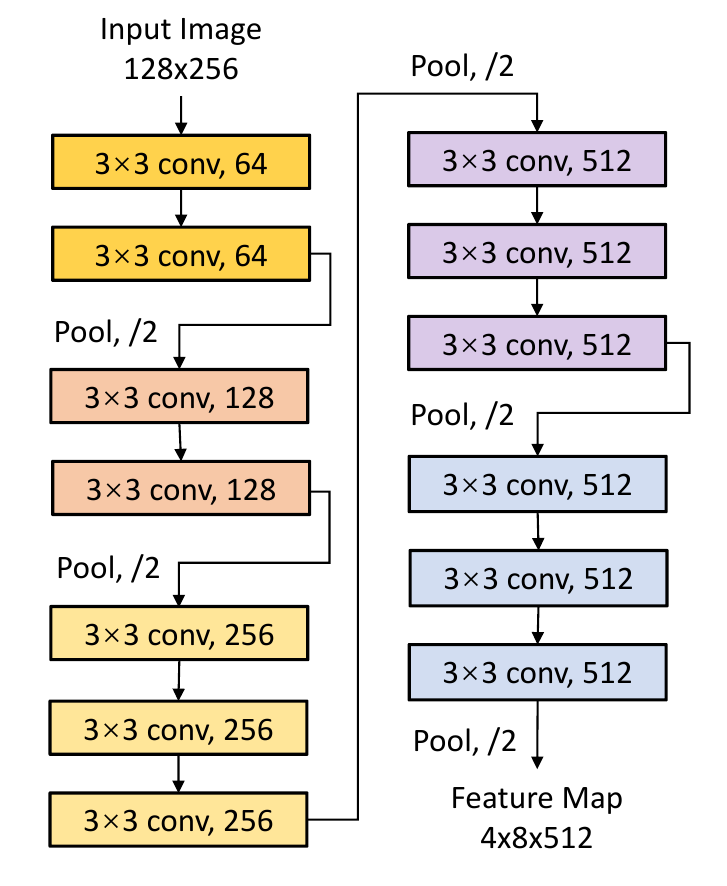}}\vspace{-0mm}
  \centerline{(b)}\medskip
\end{minipage}
 \vspace{-2mm} \caption{Encoder network in (a) UNet-ConvLSTM and (b) SegNet-ConvLSTM. Skip connections exist between convolutional layers in encoder and their matching layers in decoder.} \label{fig:encoders}
\end{figure}

Inspired by success of encoder-decoder architectures of SegNet~\cite{Badrinarayanan2017SegNetAD} and U-Net~\cite{Ronneberger2015UNetCN} in semantic segmentation, we build our network by embedding the ConvLSTM block into these two encoder-decoder networks. Accordingly, the resultant networks are named as SegNet-ConvLSTM and UNet-ConvLSTM, respectively. The encoder block and decoder block are fully convolutional networks. Thus, figuring out the number of convolutional layers and the size and number of convolutional kernels is the core of architecture design. In SegNet, the 16-layer convolution-pooling architecture of VGGNet~\cite{Simonyan2014VeryDC} is employed as its encoder.  We design the encoder by referring to SegNet and U-Net and refine it by changing the number of convolutional kernels and the overall conv-pooling layers. Thus, we can get a balance between the accuracy and the efficiency. The encoder architectures are illustrated in Fig.~\ref{fig:encoders}.

In the U-Net, a block of the encoder network includes two convolution layers with twice the number of convolution kernels as comparing to the last block and a pooling layer is used for downsampling the feature map. The size of feature map, after this operation, will be reduced to half (on side length) while the number of channels will be doubled, representing high-level semantic features. In the proposed UNet-ConvLSTM, the last block does not double the number of kernels for the convolution layers, as shown in Fig.~\ref{fig:encoders}(a). There are two reasons. First, information in original image can be well represented even with less channels. The lanes can usually be represented with primitives such as color and edge which can be well extracted and abstracted from the feature map. Second, feature maps produced by the encoder network will be fed into ConvLSTM for sequential feature learning. Parameters of a full-connection layer will be quadrupled as the side length of the feature map is reduced to half while the channels remain unchanged, which is easier to be processed by ConvLSTM. We also apply this strategy to the SegNet-ConvLSTM network.

In the decoder CNN, the size and number of feature maps should be the same as their counterparts in the encoder CNN, while arranged in an inverse direction for better feature recovering. Accordingly, the up-sampling and convolution in each sub-block of decoder match the corresponding operations in the sub-block of the encoder. In U-Net, feature-map appending is performed in a straightforward way between the corresponding sub-blocks of the encoder and decoder. While in SegNet, the deconvolution in the decoder is performed by using the indices recorded in the encoder.

For the encoder and decoder CNNs, we use the general Convolution-BatchNorm-ReLu process as a convolution operation. The `same' padding is applied by all convolutions.

\subsection{Training Strategy}
Once the end-to-end trainable neural network is built, the network can be trained to make predictions towards the ground truth by a back-propagation procedure, in which the weight parameters of the convolutional kernels and the ConvLSTM matrix will be updated. The following four aspects are considered in our training.

{\it 1)} The proposed network uses the weights of SegNet and U-Net pre-trained on ImageNet~\cite{Deng2009ImageNetAL}, a huge benchmark dataset for classification. Initializing with the pre-trained weights will not only save the training time, but also transfer proper weights to the proposed networks. When training without pre-trained weights, i.e., training from scratch, after enough epoch, the test accuracy is very close to that with pre-trained weights. Exactly, with and without pre-trained weights, the F1 values on Testset {\#}1 are 0.904 and 0.901 for `U-Net$\_$ConvLSTM', and are 0.901 and 0.894 for `SegNet$\_$ConvLSTM', respectively. This situation is in accordance with results in~\cite{He2018RethinkingIP} on object detection and in object detection and instance segmentation, but it takes more time to converge.

{\it 2)} A number of $N$ continuous images of the driving scene are used as input for identifying the lanes. Therefore, in the back propagation, the coefficient on each weight update for ConvLSTM should be divided by $N$. In our experiments, we set $N$=5 for the comparison. We also experimentally investigate how $N$ influences the performance.

{\it 3)} A loss function is constructed based on the weighted cross entropy to solve the discriminative segmentation task, which can be formulated as,
\begin{equation}
\centering
\mathcal{E}_{loss} = \sum_{\mathbf{x}\in\Omega} w(\mathbf{x})\log(p_{\ell(\mathbf{x})}(\mathbf{x})),
\end{equation}
where $\ell :\Omega\to \left\{ 1,\ldots,K \right\} $ is the true label of each pixel and $w:\Omega\to \mathbb{R}$ is
a weight for each class, on purpose of balancing the lane class. It is set as a ratio between the number of pixels in the two classes in the whole training set. The softmax is defined as,
\begin{equation}
\centering
p_k(\mathbf{x})=\exp( a_k (\mathbf{x}))/\left( \sum_{k\prime=1}^K \exp( a_k\prime (\mathbf{x})) \right),
\end{equation}
where $a_k (\mathbf{x})$ denotes the activation in feature channel $k$ at the pixel position $\mathbf{x}\in\Omega$ with $\Omega \in \mathbb{Z}^2$, $K$ is the number of classes.

{\it 4)} To train the proposed network efficiently, we used different optimizers at different training stages. In the beginning, we use adaptive moment estimation (Adam) optimizer~\cite{Kingma2014AdamAM}, which has a higher gradient descending rate but is easy to fall in local minima. It is hard to be converged because its second-order momentum is accumulated in a time window. With the change of time window, input data will also change dramatically, resulting in a turbulent learning. To avoid this, when the network is trained to a relatively high accuracy, we turn to use a stochastic gradient descent optimizer (SGD), which has a smaller meticulous stride in finding global optimal. When changing optimizer, learning rate matching should be performed, otherwise the learning will be disturbed by a totally different learning stride, leading into turbulent or stagnation in the convergence. The learning rate matching is formulated as,
\begin{equation}
\begin{split}
w_{k_{sgd}}&=w_{k_{Adam}},\\
w_{{k-1}_{sgd}}&=w_{{k-1}_{Adam}},\\
w_{k_{sgd}}&=w_{{k-1}_{sgd}}-\boldsymbol{\alpha}_{{k-1}_{sgd}}\hat{\nabla}f(w_{{k-1}_{sgd}}),
\end{split}
\end{equation}	
where $w_{k}$ denotes the weight in the $k${th} iteration, $\boldsymbol{\alpha}_{k}$ is the learning rate and $\hat{\nabla}f(\cdot)$ is the stochastic gradient computed by loss function $f(\cdot)$. In our experiments, the initial learning rate is set as $0.01$, and the optimizer is changed when the training accuracy reaches $90$\%.



\section{Experiments and results}  \label{sec:experiment}
In this section, experiments are conducted to verify the accuracy and robustness of the proposed method{\footnote{Codes are available at https://sites.google.com/site/qinzoucn}}. The performances of the proposed networks are evaluated in different scenes and are compared with diverse lane-detection methods. The influence of parameters is also analyzed.

\subsection{Datasets}

\begin{figure*}[htb]
	\centering
	\subfigure[] {
		\includegraphics[width=0.34\linewidth]{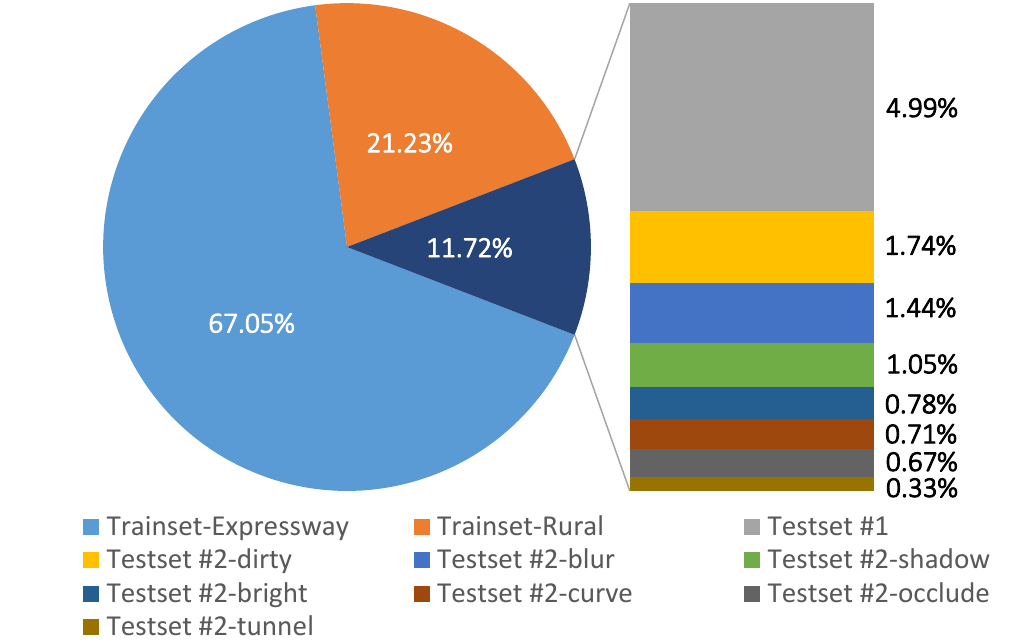}
	}
	\subfigure[]{
		\includegraphics[width=0.64\linewidth]{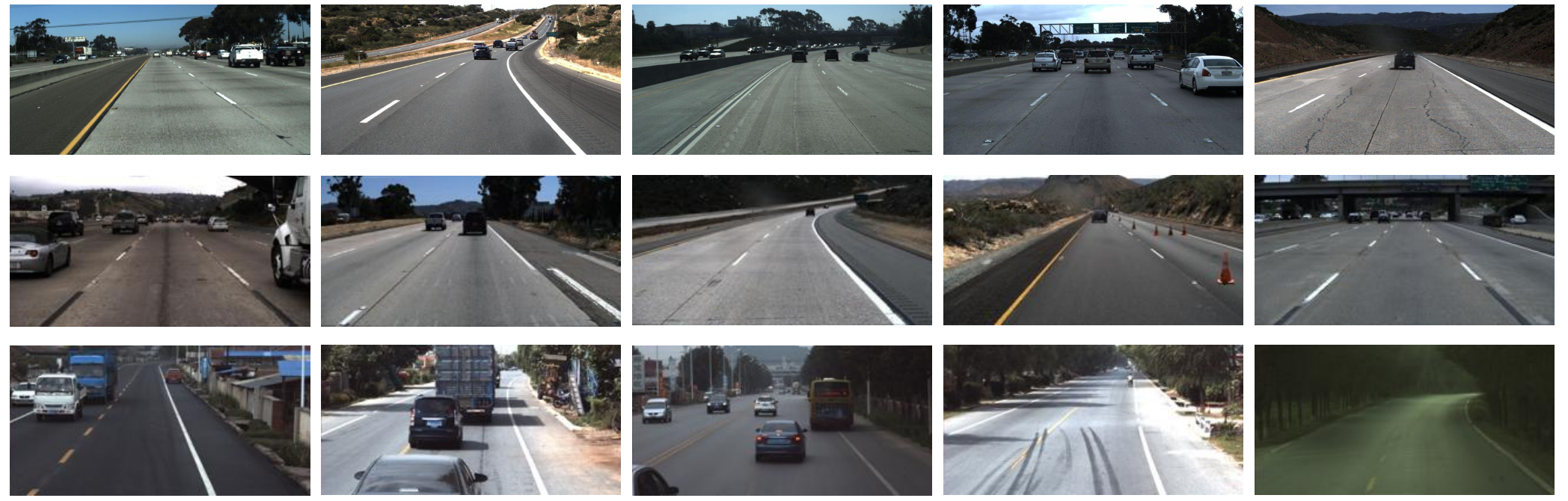}
	}
	\caption{(a) Proportion of each class in our lane-detection dataset. (b) Sample images. Top row: training images. Middle row: test images from TuSimple dataset. Bottom row: hard sample test images collected by us. }
	\label{fig:sample-data}
\end{figure*}

We construct a dataset based on the TuSimple lane dataset and our own lane dataset. The TuSimple lane dataset includes a number of 3,626 sequences of images. These images are the forehead driving scenes on the highways. Each sequence contains 20 continuous frames collected in one second. For every sequence, the last frame, i.e., 20$th$ image, is labeled with lane ground truth. To augment the dataset, we additionally labeled every 13$th$ image in each sequence. Our own lane dataset includes a number of 1,148 sequences of rural road images to the dataset. This greatly expands the diversity of the lane dataset. The detail information has been given in Table~\ref{tbl:data-training} and Fig.~\ref{fig:sample-data}.

\begin{table}[!htb]
		\caption{Construction and Content of original dataset}
        \label{tbl:data-training}
        \centering
		\setlength{\tabcolsep}{1.6mm}{
			\begin{tabular}{|l|c|c|c|}
				\hline
				\ \ Part & Including & Labeled Frames & Labeled Images \\
				\hline\hline
				Trainset & TuSimple (Highway) & 13$th$ and 20$th$ & 7,252\\
				& Ours (Rural Road) & 13$th$ and 20$th$ & 2,296\\
				\hline
				Testset & Testset {\#}1  & 13$th$ and 20$th$ & 540\\
				&Testset {\#}2 & all frames & 728\\
				\hline
		\end{tabular}}
\end{table}

For training, we sample 5 continuous images and the ground truth of the last frame as input to train the proposed network and identify lanes in the last frame. Based on the ground truth label on the 13$th$ and 20$th$ frames, we can construct the training set. Meanwhile, to fully adapted the proposed network for lane detection in different driving speeds, we sample the input images at three different strides, i.e., at an interval of 1, 2 and 3 frames. Then, there are three sampling ways for each ground-truth label, as listed in Table~\ref{tbl:data-partition}.

\begin{table}[!h]
	\renewcommand\arraystretch{1.05}
	\label{testname}
	\centering
	\caption{{\color{black}{Sampling method for continuous input images}}}\label{tbl:data-partition}
	\begin{small}
		\setlength{\tabcolsep}{1.8mm}{
			\begin{tabular}{|c|c|c|}
				\hline	
				Stride &Sampled frames& Ground Truth\\
				\hline \hline
				1& 9$th$ 10$th$ 11$th$ 12$th$ 13$th$&13$th$\\
				2& 5$th$ 7$th$ 9$th$ 11$th$ 13$th$&13$th$\\
				3& 1$th$ 4$th$ 7$th$ 10$th$ 13$th$&13$th$\\
				\hline
				1& 16$th$ 17$th$ 18$th$ 19$th$ 20$th$&20$th$\\
				2& 12$th$ 14$th$ 16$th$ 18$th$ 20$th$&20$th$\\
				3& 8$th$ 11$th$ 14$th$ 17$th$ 20$th$&20$th$\\
				\hline	
		\end{tabular}}
	\end{small}
\end{table}

In data augmentation, operations of rotation,
flip and crop are applied and a number of 19,096 sequences are produced, including 38,192 labeled
images in total for training. The input will be randomly
changed into different illumination situation, which contributes to
a more comprehensive dataset.

For test, we also sample 5 continuous images to identify lanes in the last frame, and
compare it with the ground truth of the last frame. We construct two completely different test sets - {\textbf{Testset {\#}1}} and {\textbf{Testset {\#}2}}. Testset {\#}1 is built on TuSimple test set for normal testing. Testset {\#}2 consists of hard samples collected in different situations especially for robustness evaluation.


\begin{figure}[t!]
	\centering
		\begin{minipage}[b]{0.48\linewidth}
			\centering
			\includegraphics[width=1.61in]{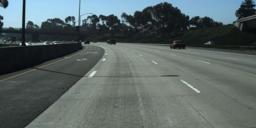}
            \centerline{(a)}\medskip
		\end{minipage}
		\begin{minipage}[b]{0.48\linewidth}
			\centering
			\includegraphics[width=1.61in]{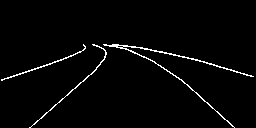}
            \centerline{(b)}\medskip
		\end{minipage}%
    \vspace{-2mm}
	\caption{An example of an input image and the labeled ground-truth lanes. (a) The input image. (b) The ground truth.}\label{fig:img-label}
\end{figure}

\begin{figure*}[t!]
	\centering
	\includegraphics[width=0.9\linewidth]{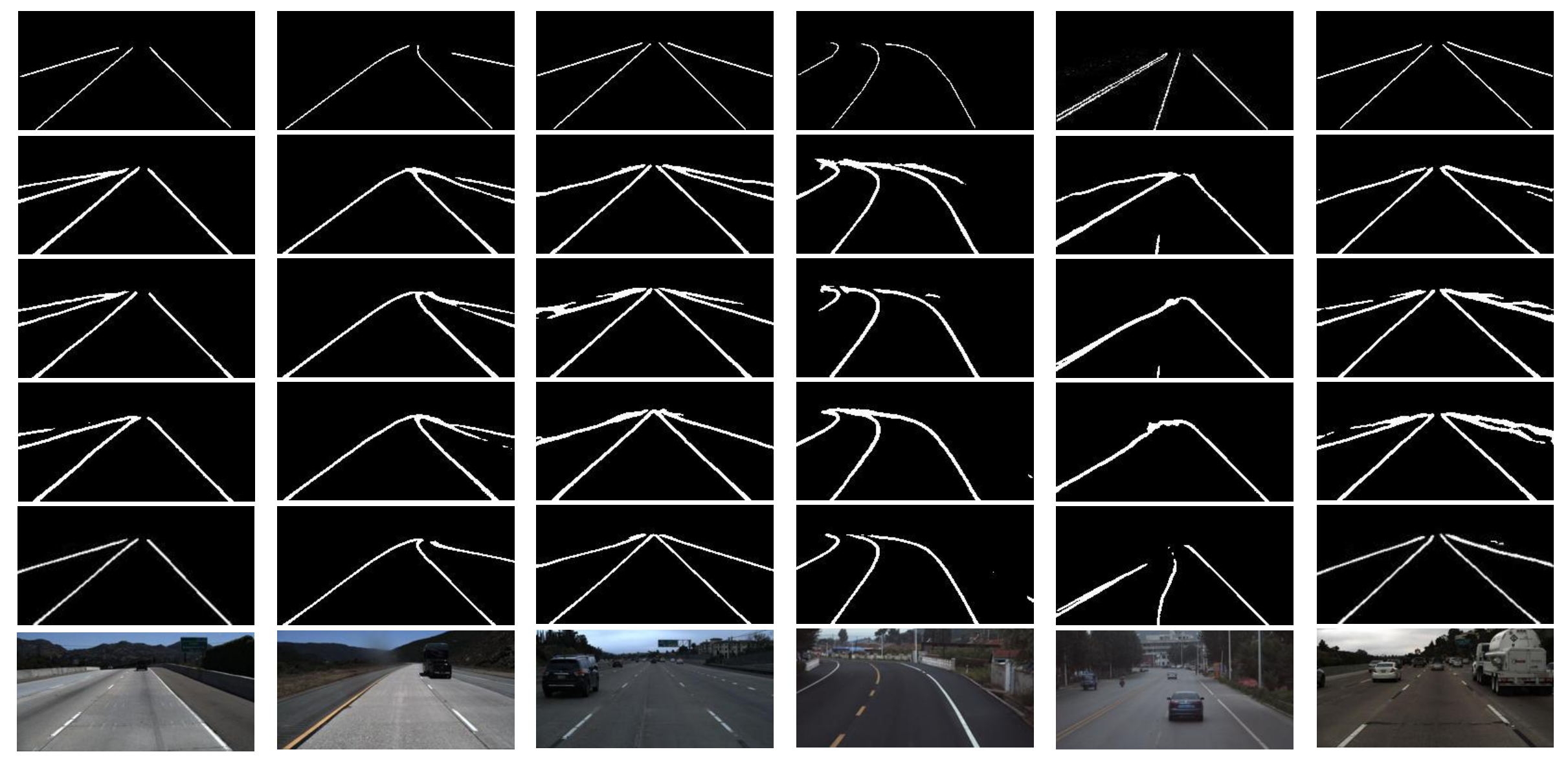}
	\caption{Visual comparison of the lane-detection results. Row 1: ground truth. Row 2: SegNet. Row 3: U-Net. Row 4: SegNet-ConvLSTM. Row 5: U-Net-ConvLSTM. Row 6: original image.}
    \label{fig:img-samples}
\end{figure*}

Due to the diversity of lane shapes, e.g., stitch-shaped lanes, single white lane and double amber lanes, a unifying standard should be established to describe these lanes in ground truth. We use thin lines to annotate the lanes. However, in the semantic segmentation task, the model have to learn from pixel-wise labels which  have refined boundaries of each object in the image. Therefore, it is not so reasonable to use thin lines
	to represent lanes since they are not semantically meaningful, for lane boundaries. A lane looks like to be more wide in a close view while much more thinner in a far view. So, we resize the input and label to lower resolution to make up for the inaccurate boundaries.
In our design, considering that the purpose of lane detection is to recognize and warn the lane departure for vehicles, it is not necessary to recognize the boundary of lanes. In the experiments, we sample the scene image into lower resolution, given the fact that lanes will become thinner when images become smaller, with a width close to one pixel. An example is shown in Fig.~\ref{fig:img-label}.
Besides, using a lower resolution can protect the model from being influenced by complex textures in the background around the vanishing point as well.

\subsection{Implementation details}
In the experiments, the images for lane detection are sampled to a resolution of 256$\times$128. The experiments for vehicle detection are implemented on a computer equipped with an Intel Core Xeon E5-2630$@$2.3GHz, 64GB RAM and two GeForce GTX TITAN-X GPUs. In order to validate the applicability of the low-resolution images and the performance of the proposed detection method, different testing conditions are used, for example the wet-, cloudy- and sunny- day scenes. In the training{\footnote{Codes are at
https://github.com/qinnzou/Robust-Lane-Detection}}, the batch size is 16, and the number of epoches is 100.

\subsection{Performance and comparison}
Two main experiments are conducted in this subsection. First, we evaluate the performance of the proposed networks, visually and quantitatively. Second, we verify the robustness of the proposed framework by testing it on difficult situations.

\subsubsection{Overall performance}
The proposed networks, i.e., UNet-ConvLSTM and SegNet-ConvLSTM, are compared with their original baselines, as well as some modified versions. To be specific, the following methods are included in the comparison.
\begin{adjustwidth}{0.5cm}{0cm}
$\bullet${\ SegNet}: It is a classical encoder-decoder architecture neural network for semantic segmentation. The encoder is the same as VGGNet.\\
$\bullet${\ SegNet}\_1$\times$1: It adds convolution with 4$\times$8 kernel between the encoder and decoder, which generates a feature map with a size of 1$\times$1$\times$512.\\
$\bullet${\ SegNet\_FcLSTM}: It uses traditional full-connection LSTM (FcLSTM) after the encoder network in SegNet\_1$\times$1, that is applying FcLSTM for extracting sequential feature from continuous images.\\
$\bullet${\ SegNet\_ConvLSTM}: It is a hybrid neural network we proposed, using convolution LSTM (ConvLSTM) after the encoder network.\\
$\bullet${\ SegNet\_3D}: It takes the continuous images as a 3D volume by laying them up. Then it uses 3D convolutional kernels to obtain mixed spatial and sequential features.\\
$\bullet${\ Five UNet-based networks}: Replacing the encoder and decoder of SegNet with that of a modified U-Net, as introduced in Section III, it generates another five networks accordingly.
\end{adjustwidth}

After training the above deep models, results obtained on the test dataset are compared. We first visually examine the outcomes obtained by different methods and then quantitatively compare them and justify the advancement of the proposed framework.

\textbf{\textit{Visual examination. }}
An excellent semantic segmentation neural network is supposed to segment the input image into different parts precisely, both in the coarse and fine level.

In the coarse level, the model is expected to predict the total number of lanes in the images correctly. To be more specific, two detection errors should be avoided in the processing of lane detection. The first one is missing detection, which predicts the true lane objects in the image as the background, and the second is excessive detection, which wrongly predicts other objects in the background as the lanes. Both these two detection errors will cause the inconsistency of number of lanes between the prediction and the ground truth, and will place a great and bad effect on the judgement of ADAS systems.

In the fine level, we wish the model to process the details precisely under the condition that the coarse target is satisfied. There should not be a great diversity on the location and length between the detected lanes and the ground truth. What's more, the severe broken of lines and fuzzy areas in the detection maps should also be avoided.

The experiment has shown that our networks achieve these two targets above and defeat other frameworks in visual examination, which can been observed in Fig.~\ref{fig:img-samples}. First, our frameworks identify each lane in the input images without missing detection or excessive detection. Other methods are easily to identify other boundaries as lane boundaries, such as the shoulder of the roadside. According to our prediction maps, every white thin line corresponds to a lane in ground truth, which shows a strong boundary distinguishing ability, ensuring a correct total number of lanes.

Second, in the outcomes of our framework, the position of lanes is strictly in accordance with the ground truth, which contributes to a more reliable and trustworthy prediction of lanes for ADAS systems in real scenes. The detected lanes of other networks are more likely to bias from the correct position with some distance. Some other methods predict an incomplete lane detection that lanes do not start from the bottom of input images and end far away from the vanishing point. In this case, the detected lane only cover a part of the ground truth. As a comparison, our method predicts the starting and ending points of lanes closer to the ground truth, presenting high detection accuracy of the length of lanes, which can be seen in the Fig.~\ref{fig:img-samples}.

Third, the lanes in our outcomes present as thin white lines, which has less blurry areas such as conglutination near vanishing point and fuzzy prediction caused by car occlusion.
Besides, our network is strong enough to identify the lanes unbrokenly when they are sheltered or have irregular shapes, avoiding cutting a continuous lane into several broken ones. These visual results show a high consistency with the ground truth.

\begin{table*}[!htp]
	\renewcommand\arraystretch{1.1}
	\centering
	\caption{Performance on Testset {\#}1.}
	\setlength{\tabcolsep}{4.1mm}{
		\begin{tabular}{|l|c|c|cc|c|c|c|c|}
			\hline
			\multirow{2}*{Methods} &
			\multicolumn{2}{c|}{Val\_Accuracy (\%)} & \multicolumn{2}{c|}{Test\_Acc (\%)} & \multirow{2}*{Precision} & \multirow{2}*{Recall}  & \multirow{2}*{F1-Measure} & \multirow{2}*{Running Time {(s)}}\\
			\cline{2-5}
			& \multicolumn{2}{c|}{Rural+Highway} &
			\multicolumn{1}{c}{Rural} & \multicolumn{1}{c|}{Highway}& & & & \\
			\hline
			\hline
			SegNet &  \multicolumn{2}{c|}{96.78}&98.16&96.93&0.796&0.962&\multicolumn{1}{c|}{0.871}&0.0052 \\
			UNet & \multicolumn{2}{c|}{96.46}&97.65&96.54&0.790&\textbf{0.985}&\multicolumn{1}{c|}{0.877}&\textbf{0.0046}\\
			
			\cline{1-9}
			SegNet\_$1\times1$ & \multicolumn{2}{c|}{94.32}&94.36&94.05&0.620&0.967&\multicolumn{1}{c|}{0.756}&0.0052\\
			UNet\_$1\times1$ & \multicolumn{2}{c|}{95.03}&95.03&94.89&0.680&0.971&\multicolumn{1}{c|}{0.800}&0.0047 \\
			\cline{1-9}
            SegNet\_FcLSTM & \multicolumn{2}{c|}{95.66}&96.11&95.82&0.723&0.966&\multicolumn{1}{c|}{0.827}& {0.0332} \\
			UNet\_FcLSTM & \multicolumn{2}{c|}{96.42}&96.71&96.33&0.782&0.979&\multicolumn{1}{c|}{0.869}& {0.0270} \\
			\cline{1-9}
            SegNet\_ConvLSTM & \multicolumn{2}{c|}{\textbf{98.52}}&98.13&97.78&0.852&0.964&\multicolumn{1}{c|}{0.901}& {0.0067} \\
			UNet\_ConvLSTM & \multicolumn{2}{c|}{98.46}&\textbf{98.43}&\textbf{98.00}&\textbf{0.857}&0.958&\multicolumn{1}{c|}{\textbf{0.904}}& {0.0058}\\
			\cline{1-9}
			SegNet\_3D & \multicolumn{2}{c|}{96.72}&96.65&96.83&0.794&0.960&\multicolumn{1}{c|}{0.868}&0.0760 \\
			UNet\_3D & \multicolumn{2}{c|}{96.56}&96.47&96.43&0.787&0.984&\multicolumn{1}{c|}{0.875}&0.0621 \\
			\hline
	\end{tabular}}
	\label{tbl:table-test1}
\end{table*}

\textbf{\textit{Quantitative analysis. }}
We examine the superior performance of the proposed methods with quantitative evaluations. The most simple evaluation criterion is the accuracy~\cite{zou2017robust}, which measures the overall classification performance based on correctly classified pixels.
{\small
\begin{equation}
\centering
\mathbf{Accuracy}=\frac{\mathbf{True\ Positive+True\ Negative}}{\mathbf{Total\ Number\ of\ Pixels}}.
\end{equation}}

As shown in Table~\ref{tbl:table-test1}, after adding ConvLSTM between encoder and decoder for sequential feature learning, the accuracy grows up about $1\%$ for UNet and $1.5\%$ for SegNet. Although the proposed models have achieved higher accuracy and previous work~\cite{Ghafoorian2018ELGANEL} also adopts accuracy as a main performance matric, we do not think it is a fair measure for our lane detection. Because lane detection is an imbalance binary classification task, where pixels stand for the lane are far less than that stands for the background and the ratio between them is generally lower than 1/50. If we classify all pixels as background, the accuracy is also about $98\%$. Thus, the accuracy can only be seen as a reference index.

Precision and recall are employed as two metrics for a more fair and reasonable comparison, which are defined as

{\small
\begin{equation}
\label{eq:pr}
\centering
\begin{split}
\mathbf{Precision}=&\frac{\mathbf{True\ Positive}}{\mathbf{True\ Positive +False\ Positive}},\\
\mathbf{Recall}=&\frac{\mathbf{True\ Positive}}{\mathbf{True\ Positive +False\ Negative}}.
\end{split}
\end{equation}	
}

In lane detection task, we set lane as positive class and background as negative class. According to Eq.~(\ref{eq:pr}), true positive means the number of lane pixels that are correctly predicted as lanes, false positive represents the number of background pixels that are wrongly predicted as lanes and false negative means the number of lane pixels that are wrongly predicted as background. As shown in Table~\ref{tbl:table-test1}, after adding ConvLSTM, there is a significant improvement in precision, and recall stays extremely close to the best results.
The precision of UNet-ConvLSTM achieves an increment of $7\%$ than the original version and its recall drops by only $3\%$. For SegNet, the precision achieves an increment of $6\%$ after adding ConvLSTM, and its recall also slightly increases.

From visual examination, we get some intuitions that the improvement in precision is mainly caused by predicting thinner lanes, reducing of fuzzy conglutination regions, and less mis-classifying. The incorrect width of lanes, the existence of fuzzy regions and mis-classifying are extremely dangerous for ADAS systems. The lanes of our methods are thinner than results of other methods, which decreases the possibility of classifying background pixels near to the ground truth as lanes, contributing to a low false positive. The reducing of fuzzy area around vanishing point and vehicle-occluded regions also results in a low false positive, because background pixels will no longer be recognized as lane class. It can be seen in Fig.~\ref{fig:img-samples}(a), mis-classifying other boundaries as lanes caused by SegNet and U-Net will be abated after adding ConvLSTM.

Relatively, the decrease in recall is also caused by the previous reasons. Thinner lanes are more accurate to represent lane locations, but sometimes it will be easy for them to keep a pixel-level distance with the ground truth. When two lines are thinner, they will be more hard to be overlapped. These deflected pixels will cause higher false negative. However, small pixel-level deviation is impalpable for human eyes, and it does not harm the ADAS systems. The reducing of conglutination has similar effect. If the model predicts all pixels in an area as lane class, which will form conglutination, where all lane pixels will be correctly classified, leading into a high recall. In this situation, even though the recall is very high, the background class will suffer from serious mis-classification and the precision will be very low.

In a word, our model fits the task better even though the recall is slightly lower. The lower recall is resultant from the thinner lines that will reasonably make slight deviation from the ground truth. Considering the precision or recall only reflect an aspect of the performance of lane detection, we introduce F1-measure as a whole matric for the evaluation. The F1 is defined as

\begin{figure*}[htb]
	\centering
	\subfigure[]{
		\includegraphics[width=0.3\linewidth]{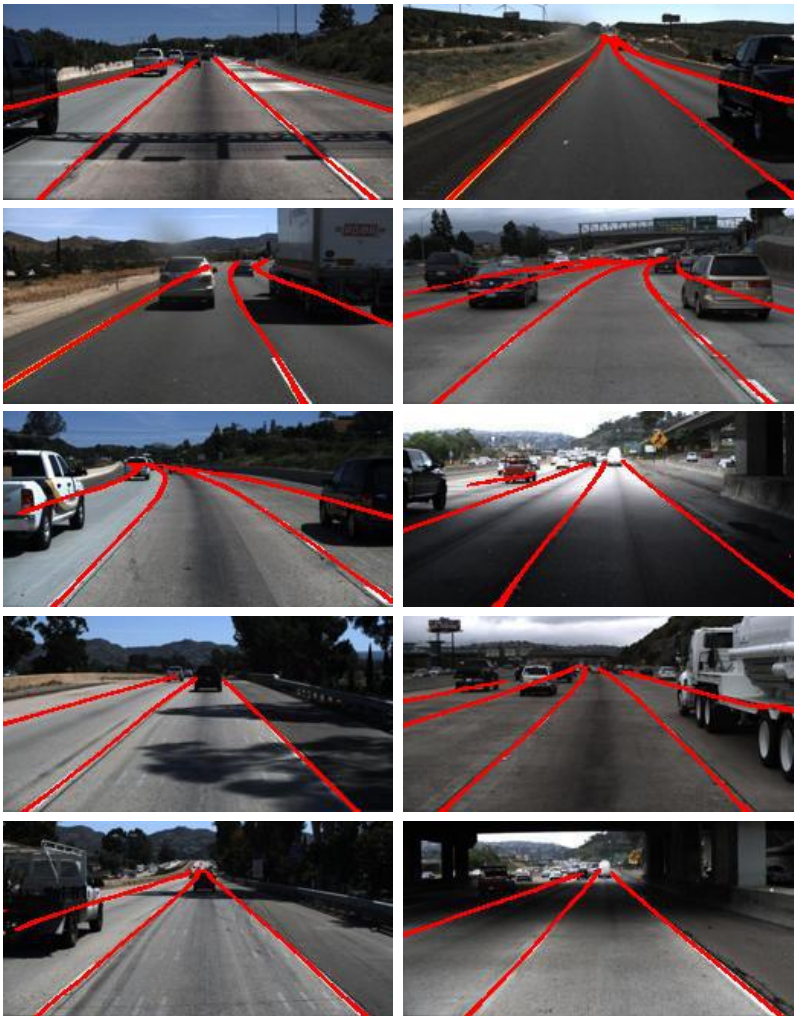}
	}
	\subfigure[]{
		\includegraphics[width=0.6\linewidth]{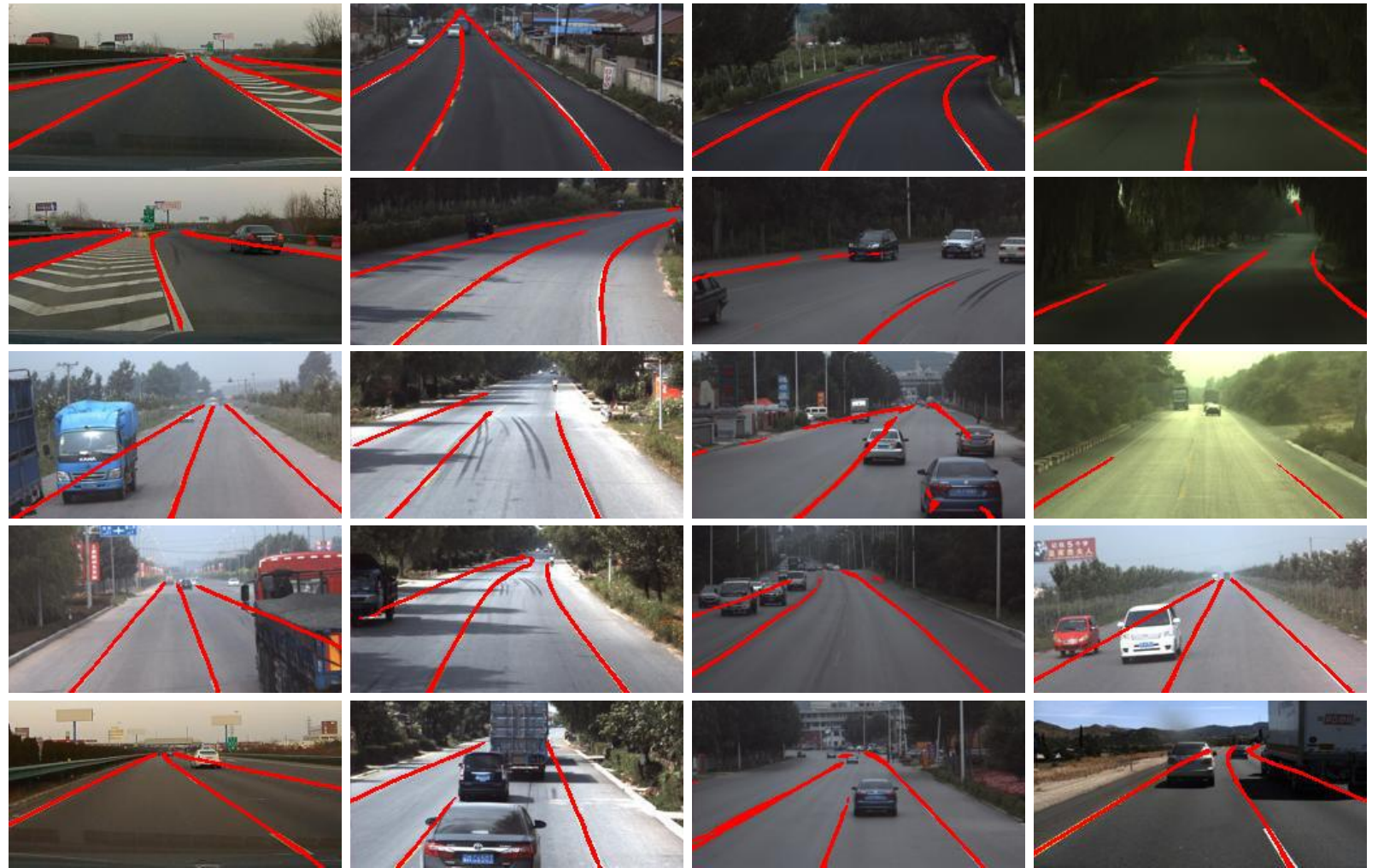}
	}
	\caption{Results obtained by UNet-ConvLSTM on challenging scenes in (a) Testset {\#}1 and (b) Testset {\#}2 without post-processing. Lanes suffered from complications of occlusion, shadow, dirty, blur, curve, tunnel and
    poor illuminance are captured in rural, urban and highway scenes by automobile data recorder at different heights, inside or outside the front windshield.}
	\label{sample}
\end{figure*}

\begin{table*}[!t]
	\centering
	\footnotesize
	\caption{TuSimple lane marking challenge leader board on test set as of March 14, 2018~\cite{Ghafoorian2018ELGANEL}.} \label{tbl:TuSimple-rank}
	\begin{tabular}{|c|c|c|c|c|c|c|}
		\hline
		Rank & Method & Name on board & Using extra data & Accuracy(\%) & FP & FN\\
		\hline \hline
		1&Unpublished&leonardoli&--&96.9&0.0442&0.0197\\
		
		2&Pan et al.~\cite{Pan2018SpatialAD}&XingangPan&True&96.5&0.0617&\textbf{0.0180}\\
		
		3&Unpublished&astarry&--&96.5&0.0851&0.0269\\
		
		4&Ghafoorian et al.~\cite{Ghafoorian2018ELGANEL}&TomTom EL-GAN&False&96.4&\textbf{0.0412}&0.0336\\
		
		5&Neven et al.~\cite{Neven2018TowardsEL}&DavyNeven&Flase&96.2&0.2358&0.0362\\
		
		6&Unpublished&li&--&96.1&0.2033&0.0387\\
		\hline

		&Pan et al.~\cite{Pan2018SpatialAD}&N/A&False&96.6&0.0609&\textbf{0.0176}\\
		&Pan et al.~\cite{Pan2018SpatialAD}&N/A&True&96.6&0.0597&0.0178\\

		&SegNet\_ConvLSTM&N/A&False&97.1&0.0437&0.0181\\	
		&SegNet\_ConvLSTM&N/A&True&97.2&0.0424&0.0184\\
		
        &UNet\_ConvLSTM&N/A&False&97.2&0.0428&0.0185\\
	
		&UNet\_ConvLSTM&N/A&True&\textbf{97.3}&0.0416&0.0186\\
		
		\hline	
	\end{tabular}
\end{table*}

{\small
\begin{equation}
\mathbf{F1}=\mathbf{2}\cdot\frac{\mathbf{Precision}\cdot \mathbf{Recall}}{\mathbf{Precision+Recall}}.
\end{equation}
}
In F1 measure, the weight of precision equals to weight of recall. It balances the antagonism by synthesizing precision and recall together without bias. As shown in Table~\ref{tbl:table-test1}, the F1-Measures of our methods rise about $3\%$ as comparing to their original versions. These significant improvements indicate that the multiple frames is better than one single frame for predicting the lanes, and the ConvLSTM is effective to process the sequential data in the semantic segmentation framework.

First, it can be seen from Table~\ref{tbl:table-test1} that, after adding FcLSTM, which are constructed on the basis of SegNet\_1$\times$1 and U-Net\_1$\times$1, the F1-Measure rises about $7\%$, indicating the feasibility and effectiveness of using LSTM for sequential feature learning on continuous images. This proves the vital role of ConvLSTM indirectly. However, performance of SegNet\_1$\times$1 and U-Net\_1$\times$1 are apparently worse than the original version of SegNet and U-Net. Due to the restriction of the input size of FcLSTM, more convolutional layers are needed to get features with 1$\times$1 size. Manifestly, the FcLSTM is not strong enough to offset lost information caused by the decreased feature size, as the accuracy it obtains is lower than the original version.
However, the accuracy of ConvLSTM versions can be improved on the basis of original baselines, due to its capability of accepting higher dimensional tensors as input.

Second, it is necessary to make comparison with other methods by taking continuous frames as inputs. 3D convolutional kernels are prevalent for stereoscopic vision problems and it also works on continuous images by piling images up to a 3D volume. The 3D convolution shows its power on other tasks such as optical-flow prediction. However, it does not perform well in lane detection task as shown in Table~\ref{tbl:table-test1}. It may be because that the 3D convolutional kernels are not so talent in describing time-series features.

To further verify the excellent performance of the proposed methods, we compare the proposed methods with more methods reported in the TuSimple lane-detection competition. Note that, our training set is constructed on the basis of TuSimple dataset. We follow the TuSimple testing standard, sparsely sampling the prediction points, which is different from pixel-level testing standard we used above. To be specific, we first map them to original image size, since the operation of crop and resize have been used in the pre-processing step in constructing our dataset. As can be seen from Table~\ref{tbl:TuSimple-rank}, our FN and FP are all close to the best results, and with a highest accuracy among all methods. Compared with these state-of-the-art methods in the TuSimple competition, the results demonstrate the effectiveness of the proposed framework.

We also train and test our networks and~Pan's method~\cite{Pan2018SpatialAD} on our dataset, with and without extra training data. When extra data are not used, these methods obtain slightly lower accuracy, higher FP and lower FN.

\begin{table*}[!t]
	\renewcommand\arraystretch{1.1}
	\centering
	\caption{{\color{black}{Performance on 12 types of challenging scenes (Top table: precision, Bottom table: F1-Measure).}}} \label{tbl:precision-f1}
	\begin{small}
		\resizebox{\textwidth}{13mm}{
			\begin{tabular}{|l|cccccccccccc|}
				\hline
				Methods&1-curve&2-shadow&3-bright&4-occlude &5-curve &6-dirty  &7-blur &8-blur &9-blur&10-shadow&11-tunnel&12-occlude\\
				\hline \hline
				{UNet-ConvLSTM}&\textbf{0.7311}&\textbf{0.8402}&\textbf{0.7743}&\textbf{0.6523}&\textbf{0.7872}&\textbf{0.5267}&\textbf{0.8297}&\textbf{0.8689}&\textbf{0.6197}&\textbf{0.8789}&\textbf{0.7969}&\textbf{0.8393}\\
				
				{SegNet-ConvLSTM}&0.6891&0.7098&0.5832&0.5319&0.7444&0.2683&0.5972&0.6751&0.4305&0.7520&0.7134&0.7648\\
				\hline
				{UNet}&0.6987&0.7519&0.6695&0.6493&0.7306&0.3799&0.7738&0.8505&0.5535&0.7836&0.7136&0.5699\\
				
				{SegNet}&0.6927&0.7094&0.6003&0.5148&0.7738&0.2444&0.6848&0.7088&0.4155&0.7471&0.6295&0.6732\\
				\hline
				{UNet-3D}&0.6003&0.5347&0.4719&0.5664&0.5853&0.2981&0.5448&0.6981&0.3475&0.5286&0.4009&0.2617\\
				
				{SegNet-3D}&0.5883&0.4903&0.3842&0.3717&0.6725&0.1357&0.4652&0.5651&0.2909&0.4987&0.3674&0.2896\\
				\hline					
		\end{tabular}}
	\end{small}
\end{table*}

\begin{table*}[!t]
	\renewcommand\arraystretch{1.1}
	\centering
	\begin{small}
		\resizebox{\textwidth}{13mm}{
			\begin{tabular}{|l|cccccccccccc|}
				\hline
				Methods&1-curve&2-shadow&3-bright&4-occlude &5-curve &6-dirty &7-blur &8-blur &9-blur&10-shadow&11-tunnel&12-occlude\\
				\hline \hline
				{UNet-ConvLSTM}&\textbf{0.8316}&\textbf{0.8952}&\textbf{0.8341}&\textbf{0.7340}&0.7826&0.3143&\textbf{0.8391}&\textbf{0.7681}&\textbf{0.5718}&0.5904&0.4076&0.4475\\
				
				{SegNet-ConvLSTM}&0.8134&0.8179&0.7050&0.6831&0.8486&0.3061&0.7031&0.6735&0.4885&\textbf{0.7681}&\textbf{0.7671}&\textbf{0.7883}\\
				\hline
				{UNet}&0.8193&0.8466&0.7935&0.7302&0.7699&\textbf{0.3510}&0.8249&0.7675&0.5669&0.6685&0.5829&0.4545\\
				
				{SegNet}&0.8138&0.7870&0.7008&0.6138&\textbf{0.8655}&0.2091&0.7554&0.7248&0.5003&0.7628&0.7048&0.5918\\
				\hline
				{UNet-3D}&0.7208&0.6349&0.6021&0.5644&0.6273&0.2980&0.6775&0.6628&0.4651&0.4782&0.3167&0.2057\\
				
				{SegNet-3D}&0.7110&0.5773&0.5011&0.4598&0.7437&0.1623&0.5994&0.6205&0.3911&0.5503&0.4375&0.3406\\
				\hline					
		\end{tabular}}
	\end{small}
\end{table*}

\begin{figure*}
	\centering
	\includegraphics[width=0.8\linewidth]{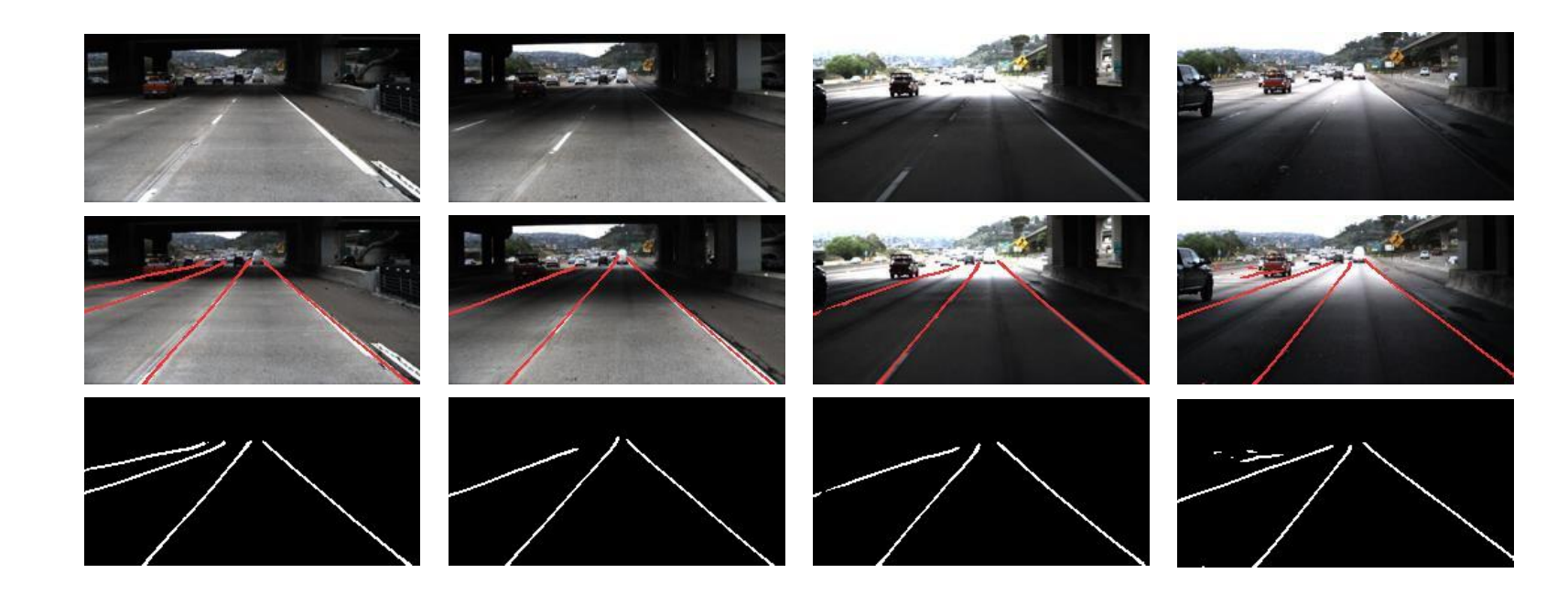}
	\caption{A hard case of lane detection when the car running into and out of shadow of a bridge. Top row: original images.  Middle row: lane predictions overlapping on the original images. Bottom row: lane predictions in white.}
	\label{fig:hard-case}
\end{figure*}

\textbf{\textit{Running Time.}}
The proposed models take a sequence of images as input and additionally add an LSTM block, it may cost more running time. It can be seen from the last column of Table~\ref{tbl:table-test1}, if processing all five frames, the proposed networks show a much more time consumption than the models that process only one single image, such as SegNet and U-Net. However, the proposed methods can be performed online, where the encoder network only need to process the current frame since the previous frames have already been abstracted, the running time will drastically reduce. As the ConvLSTM block can be executed in parallel mode with GPUs, the running time is almost the same with models that have one single image as input.

Exactly, after adding ConvLSTM block into SegNet, the running time is about 25ms if processing all five frames as new input. The running time is 6.7ms if the features of previous four frames are stored and reused, which is slightly longer than the original SegNet, whose running time is about 5.2ms. Similarly, the mean running time of U-Net-convLSTM online is about 5.8ms, slightly longer than the original U-Net's 4.6ms.

\subsubsection{Robutness}

Even though we have achieved a high performance on previous test dataset, we still need to test the robustness of the proposed lane detection model. This is because even a subtle false recognition will increase the risk of traffic accident~\cite{li2018tvt}. A good lane detection model is supposed to be able to cope with diverse driving situations such as daily driving scenes like urban road and highway, and challenging driving scenes like rural road, bad illumination and vehicle occlusion.

In the robustness testing, a brand-new dataset with diverse real driving scenes is used. The Testset {\#}2, as introduced in the dataset part, contains 728 images including lanes in rural, urban and highway scenes. This dataset is captured by data recorder at different heights, inside and outside the front windshield, and with different weather conditions. It is a comprehensive and challenging test dataset in which some lanes are hard enough to be detected, even for human eyes.

Figure~\ref{sample} shows some results of the proposed models without postprocessing. Lanes in hard situation are perfectly detected, even though the lanes are sheltered by vehicles, shadows and dirt, and in variety of illumination and road situations. In some extreme cases, e.g., the whole lanes are covered by cars and shadows, the lanes bias from road structure seams, etc., the proposed models can also identify them accurately. The proposed models also show strong adaptation for different camera positions and angles. As shown in Table~\ref{tbl:precision-f1}, UNet-ConvLSTM outperforms other methods in terms of precision for all scenes with a large margin of improvement, and achieves highest F1 values in most scenes, which indicates the advantage of the proposed models.

Since UNet-ConvLSTM outperforms SegNet-ConvLSTM in the most situations in the experiments, we recommend UNet-ConvLSTM as a lane detector in general applications. However, when strong interference is required in tunnel or occluded environments, SegNet-ConvLSTM would be a better choice.

We also test our methods with image sequences where driving enviornment change dramatically, namely, a car run into and out of shadow under bridge. Figure~\ref{fig:hard-case} shows the robustness of our method.

\begin{table*}[!t]
	\renewcommand\arraystretch{1}
	\centering
	\caption{Performance on challenging scenes under different parameter settings.}
	\footnotesize
			\begin{tabular}{|c|cccc|cccc|cccc|c|}
				\hline
				(stride, \tabincell{c}{number of \\frames} )&(3, 5)&(3, 4)&(3, 3)&(3, 2)&(2, 5)&(2, 4)&(2, 3)&(2, 2)&(1, 5)&(1, 4)&(1, 3)&(1, 2)&single \\
				\hline
				\hline
				{Total range}&12&9&6&3&8&6&4&2&4&3&2&1&--\\
				\hline
				
				{Accuracy}&0.9800&0.9795&0.9776&0.9747&0.9800&0.9795&0.9776&0.9747&0.9800&0.9795&0.9776&0.9747&0.9702\\
				{Precision}&0.8566&0.8493&0.8356&0.8171&0.8566&0.8493&0.8356&0.8171&0.8566&0.8493&0.8356&0.8171&0.7973\\
				{Recall}&0.9575&0.9582&0.9599&0.9624&0.9575&0.9582&0.9599&0.9624&0.9575&0.9582&0.9599&0.9624&0.9854\\
				{F1-Measure}&0.9042&0.8977&0.8937&0.8838&0.9043&0.8977&0.8937&0.8838&0.9043&0.8977&0.8937&0.8838&0.8821\\
				\hline					
		\end{tabular}
        \label{tbl:param_setting}
\end{table*}

\subsection{Parameter analysis}

There are mainly two parameters that may influence the performance of the proposed methods. One is the number of frames used as the input of the networks, the other is the stride for sampling. These two parameters determine the total range between the first and the last frame together.

While given more frames as the input for the proposed networks, the models can generate the prediction maps with more additional information, which may be helpful to the final results. However, in other hand, if too many previous frames are used, the outcome may be not good as lane situations in far former frames are sometimes significantly different from the current frame.
Thus, we firstly analyze the influence caused by the number of images in the input sequence. We set the number of images in the sequence from 1 to 5 and compare the results at these five different values with the sampling strides. We test it on Testset {\#}1 and the results are shown in Table~\ref{tbl:param_setting}. Note that, in top row of Table~\ref{tbl:param_setting}, the two numbers in each bracket are sampling stride and the number of frames, respectively.

It can be seen from Table~\ref{tbl:param_setting}, when using more consecutive images as input under the same sampling stride, both the accuracy and F1-Measure grow, which demonstrates the usefulness of using multiple consecutive images as input with the proposed network architecture. Meanwhile, significant improvements can be observed on the methods using multiple frames over the method using only one single image as input. While the stride length increases, the growth of performance tends to be stable, e.g., performance improvement from using four frames to five frames is smaller than that from using two frames to three frames. It may be because that the information from farther previous frames are less helpful than that from closer previous frames for lane prediction and detection.

Then, we analyze the influence of the other parameter - the sampling stride between two consecutive input images. From Table~\ref{tbl:param_setting} we can see, when the number of frames is fixed, very close performances are obtained by the proposed models at different sampling strides. Exactly, the influence of sampling stride can only be captured in the fifth decimal in the results. It simply indicates that the influence of sampling stride seems to be insignificant.

These results can also help us to understand the effectiveness of ConvLSTM. Constrained by its small size, the feature map extracted by the encoder cannot contain all the information of the lanes. So in some extent, the decoder has to imagine to predict the results. When using multiple frames as input, the ConvLSTM integrates feature maps extracted from consecutive images together, and these features enable the model to get more comprehensive and richer lane information, which can help the decoder to make more accurate predictions.

\section{Conclusion} \label{sec:conclusion}
In this paper, a novel hybrid neural network combining CNN and RNN was proposed for robust lane detection in driving scenes. The proposed network architecture was built on an encoder-decoder framework, which takes multiple continuous frames as an input, and predicts the lane of the current frame in a semantic segmentation manner. In this framework, features on each frame of the input were firstly abstracted by a CNN encoder. Then, the  sequential encoded features of all input frames were processed by a ConvLSTM. Finally, outputs of the ConvLSTM were fed into the CNN decoder for information reconstruction and lane prediction. Two datasets containing continuous driving images are constructed for performance evaluation.

Compared with baseline architectures which use one single image as input, the proposed architecture achieved significantly better results, which verified the effectiveness of using multiple continuous frames as input. Meanwhile, the experimental results demonstrated the advantages of ConvLSTM over FcLSTM in sequential feature learning and target-information prediction in the context of lane detection. Compared with other models, the proposed models showed higher performance with relatively higher precision, recall, and accuracy values. In addition, the proposed models were tested on a dataset with very challenging driving scenes to check the robustness. The results showed that the proposed models can stably detect the lanes in diverse situations and can well avoid false recognitions. In parameter analysis, longer sequence of inputs were found to make improvement on the performance, which further justified the strategy that multiple frames are more helpful than one single image for lane detection.

In the future, we plan to further enhance the lane detection system by adding lane fitting into the proposed framework. In this way, the detected lanes will be smoother and have better integrity.
Besides, when strong interferences exist in dim environment, SegNet-ConvLSTM was found to function better than UNet-ConvLSTM, which needs more investigations.


\bibliographystyle{IEEEtran}
\bibliography{refs}

\end{document}